\documentclass[runningheads]{llncs}
\usepackage{svg}

\newcommand{\orcid}[1]{\href{https://orcid.org/#1}{\includesvg[height = 2ex]{ORCIDiD_iconvector}}}

\usepackage{url}

\usepackage{cite}
\usepackage{amsmath,amssymb,amsfonts}
\usepackage{graphicx}
\usepackage{textcomp}
\usepackage{xcolor}
\usepackage{algorithm}
\usepackage[]{algpseudocode}
\usepackage{subcaption}
\usepackage{float}
\usepackage{url}
\usepackage{tikz}
\usepackage{multirow}
\usepackage{hyperref}
\usepackage{arydshln}

\usepackage{colortbl}

\def\BibTeX{{\rm B\kern-.05em{\sc i\kern-.025em b}\kern-.08em
    T\kern-.1667em\lower.7ex\hbox{E}\kern-.125emX}}
    
\usepackage{flushend}

\begin{document}
\mainmatter              % start of a contribution
%
% \title{Visualisation of Multi-Objective Search Trajectories}
\title{Search Trajectories Networks of Multiobjective Evolutionary Algorithms}
\titlerunning{STNs on MOPs}  % abbreviated title (for running head)
\author{
Yuri Lavinas\inst{1} \and 
Claus Aranha\inst{2} \and 
Gabriela Ochoa\inst{3}
}

% \author{First Author\inst{1}\orcid{0000-1111-2222-3333} \and
% Second Author\inst{2,3}\orcid{1111-2222-3333-4444} \and
% Third Author\inst{3}\orcid{2222--3333-4444-5555}}

\authorrunning{Y. Lavinas, C. Aranha, G. Ochoa} % abbreviated author list (for running head)
% \tocauthor{Ivar Ekeland, Roger Temam, Jeffrey Dean, David Grove,
% Craig Chambers, Kim B. Bruce, and Elisa Bertino}
%
\institute{University of Tsukuba, Tsukuba, Japan,\\
\email{lavinas.yuri.xp@alumni.tsukuba.ac.jp}
\and
University of Tsukuba, Tsukuba, Japan,\\
\email{caranha@cs.tsukuba.ac.jp}
\and
University of Stirling,Stirling, UK \\
\email{gabriela.ochoa@cs.stir.ac.uk}
} 

\widowpenalty=1000
\clubpenalty=1000

\maketitle              % typeset the title of the contribution

\begin{abstract}

Understanding the search dynamics of multiobjective evolutionary algorithms (MOEAs) is still an open problem.  
This paper extends a recent network-based tool, search trajectory networks (STNs), to model the behavior of MOEAs. Our approach uses the idea of decomposition, where a multiobjective problem is transformed into several single-objective problems.
We show that STNs can be used to model and distinguish the search behavior of two popular multiobjective algorithms, MOEA/D and NSGA-II,  using 10 continuous benchmark problems with 2 and 3 objectives. Our findings suggest that we can improve our understanding of MOEAs using STNs for algorithm analysis.

\end{abstract}

\section{Introduction}

Most real-world optimization problems involve multiple conflicting objectives. This has prompted the development of a variety of multiobjective evolutionary algorithms (MOEAs), which can be classified into three broad categories, based on dominance \cite{deb2002fast}, indicators \cite{beume2007sms} and decomposition \cite{zhang2007moea}. There has been significant progress in improving MOEAs in all categories, and these algorithms are widely used in practice. However, algorithm development and improvement are mostly guided by intuition and empirical performance comparisons. We argue that there is a lack of accessible tools to analyze, contrast and visualize the dynamic behavior of MOEAs.

The main goal of this article is to generalize to multiobjective optimization a recent graph-based modeling tool, search trajectory networks (STNs), which was originally proposed for single objective optimization \cite{stn_evostar,stn_main}. Our approach uses decomposition, a key strategy in multiobjective optimization, where the multiobjective problem is transformed into several single-objective problems. This is convenient as it allows us to use the existing tools and features proposed for single objective STNs \cite{stn_evostar,stn_main}.

This study is exploratory and interpretative in nature, taking the form of a case study using 10 continuous benchmark problems with 2 and 3 objectives and two broadly known MOEAs, MOEA/D~\cite{zhang2007moea} and NSGA-II~\cite{deb2002fast}.  To the best of our knowledge, this is the first effort to apply STNs for modeling the search behavior of MOEAs.

The paper is organized as follows. Section \ref{related-work} overviews previous work related to visualization in multiobjective optimization. Section \ref{prelim} introduces relevant concepts. Our proposal to extend STNs to model MOEAs is described in Section \ref{mo-stns}. The experimental setup and results are presented in Sections \ref{experiments} and \ref{results}, respectively. Finally, Section \ref{conclusions} outlines our main findings and suggestions for future work.
\section{Related Work}
\label{related-work}

Most visualization approaches in the literature for multiobjective optimization focus entirely on the objective space, completely ignoring the decision space. The classical visualization shows the true or approximated Pareto front for 2 or 3 objectives in a standard scatter plot.   Extensions to this idea for visualizing problems with more than three objectives, using dimensionality reduction techniques, have been proposed \cite{Tusa2015}. However, by purely focusing on the objective space, the interaction effects from the decision variables are ignored; therefore, almost no information on the structural properties of the problem landscapes can be derived.

Very few visualization techniques in the literature provide a joint view of the decision and objective spaces in continuous multiobjective optimization. Fonseca et al. \cite{performance_assessment} proposed the cost landscapes, which use dominance ranking to evaluate points in the decision space with respect to global optimal trade-offs. This approach, however, does not capture local optimal sets.  Kerschke and Grimme \cite{KerschkeG17}  proposed the gradient field maps to explicitly address local optimal sets, with further extensions plotting landscapes with optimal trade-offs \cite{OnePLOTtoShow}, and providing an accessible dashboard for visualization \cite{SchapermeierGK21}. In combinatorial optimization, recent work has adapted the local optima networks model \cite{LON2008} to multiobjective optimization, providing visual insights into the distribution and connectivity pattern of Pareto local optimal solutions \cite{Liefooghe2018OnPL} and dominance-based hill-climbing \cite{Fieldsend2019VisualisingTL}. These are insightful visual approaches; however, they concentrate on the structural configuration of local and global optima in the optimization landscapes, rather than on the dynamic (trajectory) behavior of the search process.

\section{Preliminaries}
\label{prelim}
\subsection{Search Trajectory Networks}

The original STN model definitions for single objective optimization can be found in~\cite{stn_main}; we rephrase them here for completeness and to guide our proposed extension to multiobjective optimization. To define a network model, we need to specify their nodes and edges. The relevant definitions are given below.
\begin{description}
 
\item{\bf Representative solution.} Is a solution to the problem at a given iteration that represents the status of the search process. For population-based algorithms, the solution with the best fitness in the population at a given generation is chosen as the representative solution.

\item[Location.] Is a non-empty subset of solutions that results from a mapping process. Each solution in the search space is mapped to one location. Several similar solutions are generally mapped to the same location, as the locations represent a partition of the search space.  

\item[Search trajectory.] Given a sequence of representative solutions in the order in which they are encountered during the search process, a search trajectory is defined as a sequence of locations formed by replacing each solution with its corresponding location. 

\item[Node.] Is a  location in a search trajectory of the search process being modeled.  The set of nodes is denoted by $N$.

\item[Edges.] Edges are directed and connect two consecutive locations in the search trajectory. Edges are weighted with the number of times a transition between two given nodes occurred during the process of sampling and constructing the STN. The set of edges is denoted by $E$. 

\item[Search trajectory network (STN).] Is a directed graph $\mathrm{STN} = (N,E)$, with node set $N$, and edge set $E$ as defined above.

\end{description}

The data to construct an STN model is gathered while the studied algorithm is running. Specifically, the required output from a run is a list of steps connecting two adjacent representative solutions in the search process. Each search step is stored as an entry in a log file containing the two consecutive representative solutions being linked with the step; these transitions become the edges of the network model. Once the data logs of a predefined number of runs of a given algorithm-problem pair are gathered, a post-processing maps solutions to locations, aggregates all the locations and transitions, and constructs a network object.

\subsection{Multiobjective Optimisation Problems}

In Multiobjective Optimization Problems (MOPs), a solution to the problem is evaluated by multiple objective functions, which are possibly conflicting. One key characteristic of MOPs is that a single solution rarely is the optimal solution for all objectives. This leads to the concept of ``Pareto dominance'': A solution $x_j$ is said to be dominated by $x_i$ if $x_i$ is better than $x_j$ in one objective, and at least equivalent in all others. Otherwise, $x_j$ is non-dominated by $x_i$. Note that two solutions can be mutually non-dominated (each solution is better in a different objective).

%%%% Yuri's original text %%%%%%%%
% Multiobjective optimization problems (MOPs) are problems with multiple, conflicting objectives that are optimized simultaneously. Consequently, the goal of the MOP optimization algorithm is to find the approximate set of solutions that optimally balance the different objectives.

%A critical concept in the MOPs domain is the ``Pareto dominance''. A solution $X$ dominates solution $Y$ if solution $X$ has a better trade-off in at least one of the objectives of a given MOP. On the contrary, a non-dominated solution exists if no other solution provides a better trade-off in all objectives. Then, the set of all Pareto-Optimal solutions forms the Pareto-optimal Set (PS), and their image set is referred to as the Pareto-Optimal Front (PF).
%%%%%%%%%%%%%%%%%%%%%%%%%%%%%%%%%%

As a consequence of Pareto dominance, the search status of an MOEA cannot be represented by a single representative solution. Therefore, the definitions of location, node, edge, and search trajectory must be extended in a multiobjective context. Our proposed extension of STNs to multiobjective optimization relies on a few operations that are used in dominance-based MOEAs such as NSGA-II~\cite{deb2002fast} and decomposition-based MOEAs such as MOEA/D~\cite{zhang2007moea}. Brief descriptions of these operations are given below.

\begin{description}
 
\item[\bf Non-dominated sorting.] Sorts a set $X$ of solutions based on their dominance relationship. Initially, we find the subset of solutions, $X_{ND}$, in $X$ that are non-dominated in relation to all other solutions and give the solutions in $X_{ND}$ rank 0. Then we repeat this procedure on the set $X - X_{ND}$, giving the new non-dominated solution set the rank 1. The procedure keeps repeating until we have ranked all solutions in $X$.

% \item[\bf Non-dominated sorting.] Orders the set of solutions based on the non-dominance relationship. The better ranks are given to the non-dominated solutions, defining the first front. The second front has solutions that are dominated only by the solutions from the first front. That is, if we exclude the first-front solutions, solutions of the second-front solutions become non-dominated solutions. This idea applies to all fronts. Thus solutions of a front are dominated only by solutions of the previous fronts and are non-dominated among each other. 

\item[\bf Decomposition.] Breaks down a MOP into a set of single objective subproblems, where each subproblem is a combination of the objectives, characterized by a weight vector. There are several methods to generate a set of weight vectors (decomposition methods). In this work, we use the Uniform Design~\cite{uniform_design}, as it allows us to choose the number of weight vectors to be generated explicitly.

% \item[\bf The decomposition technique.] The idea of the vector decomposition technique is to decompose the MOP into a given number of scalar single-objective subproblems using weight vectors. Use chose to use the Uniform Design decomposition method~\cite{uniform_design} because it allows us to decide the number of weight vectors explicitly.

\item[\bf Scalar aggregation function.] Takes the objective values of one solution and the weight vector of one subproblem and calculates a scalar value that represents the quality of that solution for that particular subproblem. In a decomposition-based MOEA, these quality values are used to associate one representative solution to each subproblem. There are several scalar aggregation functions (also called scalarization functions), and in this work, we use the Weighted Tchebycheff~\cite{miettinen1999nonlinear, moeadr_paper}, which is less affected by the shape of the Pareto front in a given MOP.

% \item[\bf Scalar aggregation function.] Is a function that measures the scalar value of candidate solutions for each of the subproblems defined by the weight vectors. For each subproblem, we can calculate the scalar values of solutions and compare these values easily. Moreover, these methods associate only one candidate solution to each one of the vectors. We chose to use the Weighted Tchebycheff function~\cite{miettinen1999nonlinear, moeadr_paper} because this function is not affected by the features of the Pareto front.

\end{description}
\section{STN Extension for the Multiobjective Domain}
\label{mo-stns}

As discussed in the previous section, STNs are calculated based on the sequence of representative solutions in an algorithm execution. However, each iteration in an MOEA does not contain a single representative solution. In this section, we describe a method to calculate STNs for multiobjective optimization.

The key idea of this method is that we keep track of a small number of decomposition vectors, match a representative solution to a vector, and then merge the vector trajectories of each vector into a single multiobjective STN. The following processes describes the steps to associate a solution to each of the vectors:

% As previously discussed, STNs are especially good for comparing algorithms because they can model the search trajectories. Here, we describe the main steps in the process of modeling the STNs for MOPs.

% Since the STN models depend on the connection of consecutive best solutions, one of the most critical steps for creating an STN is to choose the representative solution at each iteration. However, in MOPs, there is no single best solution but a set of them. Therefore, the main challenge in creating STNs in MOPs is defining a process to select these best solutions. The following steps describe the process we use in this current work to select the best solutions for the multiobjective STN:

%%%%%%%%%%%%%%%%%%%%%%

\begin{enumerate}
    \item Initially, we choose the number of decomposition vectors $n$ and generate these vectors using the decomposition technique, using the technique discussed in the previous section.
    \item Every iteration, we apply the non-dominated sorting procedure on the solution set and select the set of rank 0 solutions as the set of candidate solutions. If the number of solutions in this set is less than $n$, we add solutions from rank 1, 2, and so on until we have $n$ or more candidate solutions.
    \item Every iteration, we select a representative solution for each vector from the set of candidate solutions, using the scalar aggregation function. In the case of ties, we choose the newest solution, discarding solutions already associated with that vector.
\end{enumerate}

% \begin{enumerate}
%     \item Define how many decomposition vectors to use for modeling, $n$, using the decomposition technique.
%     \item Rank the candidate best solutions of a generation using the non-dominated sorting. We accept as candidate solutions all solutions in the first front, the non-dominated solutions. If the number of solutions in the first front is less than $n$, we accept solutions from the second front until we have $n$ or more candidate solutions.
%     \item  Match the candidate best solutions of a generation to each trajectory vector, using the scalar aggregation function. For minimization problems, smaller scalar values are preferred, and in the case of ties, we accept the newest solution and discard the solution already associated with a vector.
% \end{enumerate}

After associating one representative solution for each weight vector, we map the locations of each representative solution as the nodes. To create this mapping, we use a precision parameter to portion the space into hypercubes with length $10^{-03}$. In other words, a node in the STN is equivalent to a $0.001 \times 0.001 \times 0.001$ hypercube. Edges are given by the sequence of locations of consecutive iteration. Each decomposition vector is modeled as an STN. Finally, we merge these STNs, establishing the merged STN, that models the search progress of an algorithm in a problem. We merge the STNs as follows:

\begin{enumerate}
\item The merged STN model merges the $n$ STNs of each decomposition vector and is obtained by the graph union of the $n$ individual graphs.
\item The merged graph contains the nodes and edges present in at least one of the vectors graphs. Attributes are kept for the nodes and edges, indicating whether they were visited by both algorithms (shared) or by one of them only.
\end{enumerate}

\section{Experiments}
\label{experiments}

We conduct the following experimental and exploratory study to investigate whether STNs can discriminate between MOEA/D-DE and NSGA-II. This analysis is conducted by characterizing and visualizing the search behavior of these MOEAs. 

% \subsection{Benchmark Problems}
We use the UF benchmark set~\cite{zhang2008multiobjective}, with 2 and 3 objectives. Both algorithms were widely studied, and collectively these works have shown that MOEA/D performs better than NSGA-II in this group of problems. Thus the UF benchmark set helps show that the STN can discriminate these metaheuristics. For all functions in all benchmark sets, we set the number of dimensions to $D = 10$. The implementation of the test problems available from the \textit{smoof} package~\cite{smoof} was used in all experiments. The UF Benchmark set comprises ten unconstrained test problems with Pareto sets designed to be challenging to existing algorithms~\cite{li2019comparison}. Problems UF1-UF7 are two-objective MOPs, while UF8-UF10 are three-objective problems~\cite{zhang2008multiobjective}.

\subsection{Experimental Parameters}~\label{parameters}

\begin{table}[htbp]
\centering
	\small
	\caption{Experimental parameter settings.}
	\label{chap4:parameter}
    \begin{tabular}{l|l}
        % \hline
        \rowcolor[gray]{.85}MOEA/D Parameters             & Value            \\ \hline
        DE mutation                                  & $F = 0.25$       \\ \hline
        \multirow{2}{*}{Polynomial mutation}        & $\eta_m = 20$    \\
                                                            & $prob = 0.01$     \\ \hline
        \multirow{2}{*}{Restricted Update}                            & $nr = 2$         \\ %\hline
                                          & $\delta_p = 0.9$ \\ \hline
        Neighborhood size                                   & $T = 20\%$ of the pop. size         \\ %\hline
        % \multicolumn{2}{c}{} \\
        \hline
        SLD weight vectors & 250  \\ %\hline
                                                     
        \multicolumn{2}{c}{}\\ %\hline
        \rowcolor[gray]{.85}NSGA-II Parameters             & Value            \\ \hline
        Tournament Size                                            & 2 \\ \hline
        % \multirow{2}{*}{SBX crossover}        & $\eta = 20$    \\
        %                                                     & $prob = 0.9$     \\ \hline
        DE mutation                                  & $F = 0.25$       \\ \hline
        \multirow{2}{*}{Polynomial mutation}        & $\eta_m = 3$    \\
                                                            & $prob = 0.1$     \\ %\hline

        \multicolumn{2}{c}{}\\ %\hline
        
        \rowcolor[gray]{.85} Population size  & Value            \\ \hline
        MOEA/D  and NSGA-II                                 & 250     \\ %\hline    
        \multicolumn{2}{c}{}\\ %\hline 
        \rowcolor[gray]{.85}Experiment Parameters           & Value            \\ %\hline
        Repeated runs                                       & 3               \\ \hline
        Computational budget                                & 30000 evals.            \\ %\hline
\end{tabular}
\end{table}

We used the MOEA/D variant using the Differential Evolution (MOEA/D-DE) parameters as they were introduced in the work of Li and Zhang~\cite{li2009multiobjective} in all tests. Details of these parameters can be found in the documentation of package MOEADr and the original MOEA/D-DE reference~\cite{zhang2009performance,moeadr_package,moeadr_paper}. We used the NSGA-II parameters as they were introduced in the work of Deb et al.~\cite{deb2002fast} in all tests, except for the use of the DE mutation operator, with the same parameters as MOEA/D.  Details of these parameters can be found in the documentation of package nsga2R and the original reference~\cite{nsga2R_package}. Table (\ref{chap4:parameter}) summarizes the experimental parameters for both algorithms.

% Visual argument of more than 5 is not manageable - 1st work
% For creating the STNs, we use $n = 5$ different vectors, and for each vector, there is one solution at a given iteration. We understand that the choice of the weight vector distribution is of critical importance. That said, our idea here was to use a small number of vectors that could capture visually valuable information about the pair algorithm-problem. In a recent similar work on the combinatorial domain, Cosson et al.~\cite{Cosson2021DecompositionBasedML} suggests that using values small values for the number of vectors are a good choice for two objective MOPs. Thus, we decide to use $5$ as a suitable choice for two and three objectives problems.

For creating the STNs, we use $n = 5$ different vectors, and for each vector, there is one solution at a given iteration. Our idea here was to use a small number of vectors that could capture visually valuable information about the pair algorithm-problem. More work is required to understand the effect of different values for the number of vectors. %In a recent similar work on the combinatorial domain, Cosson et al.~\cite{Cosson2021DecompositionBasedML} suggests that using values small values for the number of vectors are a good choice for two objective MOPs. Thus, we decide to use $5$ as a suitable choice for two and three objectives problems.

% In this work, we create $5$ different components, each of one best solutions at a given iteration. We understand that the choice of the weight vector distribution is of critical importance. That said, our idea here was to use a small number of components that could capture valuable information about the pair algorithm-problem. In a recent similar work on the combinatorial domain, Cosson et al.~\cite{Cosson2021DecompositionBasedML} suggests that using values small values are a good choice for two objective MOPs. However, values of $1$, $2$, or $3$ are too small since the first one generates a single vector that reduces the problems to only a single area of the objective space. In contrast, the other two values generate boundaries vectors that consider the MOP single-objective for the two-objective and the three objective cases, respectively.~\footnote{In other words, using $2$ in a two objective MOP or $3$ in a three objective MOP leads to a focus on only one objective for each of the vectors.} Thus, we decide to use $5$ as a suitable choice for both two and three objectives problems, although more work needs to be done to better understand the true impact of the choice of parameter. %With $s=5$ and using the Uniform Design decomposition~\cite{uniform_design}, for the 2 objectives MOPs the vectors are: V1 (0.9, 0.1), V2 (0.7, 0.3), V3 (0.5,0.5), V4 (0.3, 0.7) and V5 (0.1, 0.9). For the 3 objectives MOPs, the vectors are: V1 (0.68, 0.22, 0.09), V2 (0.45, 0.16, 0.38), V3 (0.29, 0.64, 0.07), V4 (0.16, 0.42, 0.42) and V5 (0.05, 0.09, 0.85).

\subsection{Metrics}~\label{evaluation_traditional}

\paragraph{MOP metrics:} We use the following criteria to compare the results of the different strategies: (a) final approximation hypervolume (HV) and Inverted Generational Distance (IGD). For calculating the hypervolume, we set the reference point to (1.1,1.1) for two objective problems and (1.1,1.1,1.1) for three objective problems.

% \textit{HV:} the hypervolume can be defined as the volume of the n-dimensional polygon formed by some reference point, $\vec{x}_{\text{ref}} = (x_1, x_2, ..., x_m)$ ,  and a finite set $S = (x_1, x_2, ..., x_m)$ of solutions in the positive orthant, $\mathbb{R}^{n}_{\geq 0}$, with $m$ being the number of objectives~\cite{beume2009complexity}.

% \textit{IGD:} the Inverted Generational Distance (IGD) as the average distance from each reference point to its nearest solution(Ishibuchi et al.~\cite{ishibuchi2015modified}). Mathematically IGD can be defined as:

% \begin{equation}
% \centering
%     IGD(PF) = \frac{1}{|Z|} \sum\limits_{i=1}^{|Z|} \hat{d}_i,
% \end{equation}

% where $PF = (x_1, x_2, ..., x_m)$, $m$ is the number of objectives, is a finite reference set of solutions in the positive orthant, $\mathbb{R}^{n}_{\geq 0}$, $Z = \{z_1, z_2,..., z_{|z|}\} $, is the reference set of solutions of size $n$, and $\hat{d}$ is the Euclidean distance from $z_j$, $j \in 1,..., n$ to its nearest objective solution in $PF$. Small IGD values suggest that the set of solutions evaluated have both good convergence and diversity, provided with a well-distributed reference set of solutions~\cite{ishibuchi2015modified}.

\paragraph{STN metrics:} We use seven STN metrics to assess the global structure of the trajectories and bring insight into the behavior of the MOEAs modeled. These metrics are (1) the number of unique nodes and (2) the number of unique edges, and we also calculate (3) the number of shared nodes between vectors, (4) the number of Pareto optimal solutions, (5) the number of components of the network and (6) the mean and (7) maximum values of the in- and out-degree of nodes. These metrics are summarised in Table~\ref{stn_metrics}. It is worth noting that additional metrics could also be considered, such as the distances between optimal solutions and centrality metrics for the optimal solutions.

\begin{table}[htbp]
\centering
\caption{Description of STN metrics}
\label{stn_metrics}
\begin{tabular}{c|c}
\rowcolor[gray]{.8} Metric & Justification \\ \hline
    Number of nodes & Shows unique locations visited.\\\hline
    Number of edges & Shows unique search transitions between nodes.\\\hline
    Number of shared nodes & Shows nodes visited by more than one components.\\\hline
    Number of optimal solutions & Shows what nodes are in the theoretical PF.\\\hline
    Number of Components & Show the distribution of the search progress.\\\hline
    Mean in/out-degree & Shows the mean degree of branching and loops. \\\hline
    Maximum in/out-degree & Shows the maximum degree of branching and loops.\\%\hline
\end{tabular}
\end{table}

\subsection{Reproducibility}

For reproducibility purposes, all the code and experimental scripts are available online at \href{https://github.com/gabro8a/STNs-MOEA.git}{https://github.com/gabro8a/STNs-MOEA.git}. 

\section{Results}
\label{results}

\begin{table}[t]
\centering
\caption{\textit{HV}, \textit{IGD}, \textit{Nodes}, Ratio of \textit{Edges} given the number of nodes, Ratio of \textit{Shared} Nodes given the number of nodes, Number of \textit{Opt}imal Solutions, \textit{Comp}onents, \textit{Max} number of \textit{in}-degrees, \textit{Mean} number of \textit{in}-degrees, \textit{Max} number of \textit{out}-degrees, \textit{Mean} number of \textit{out}-degrees of MOEA/D and NSGA-II.}
\label{mop_metrics}
\begin{tabular}{c|cc|ccccccccc}
    % \hline
    \rowcolor[gray]{.8}\multicolumn{12}{c}{\textbf{MOEA/D-DE}}\\\hline
    \rowcolor[gray]{.95} & HV & IGD & Nodes &  Edges & Shared & Opt.  & Comp. & Max in. & Mean in. & Max out. & Mean out. \\ 
    \hline
    UF1 & 0.86 & 0.01 & 1222& 1.06 & 0.01 & 0 & 6  & 4  & 1.06 & 4 & 1.06 \\
    UF2 & 0.86 & 0.01 & 1422& 1.04 & 0.01 & 0 & 10 & 3  & 1.04 & 3 & 1.04 \\
    UF3 & 0.57 & 0.17 & 940 & 1.12 & 0.20 & 4 &  1 & 15 & 1.12 & 17& 1.12 \\
    UF4 & 0.47 & 0.04 & 937 & 1.10 & 0.00 & 0 & 15 & 4  & 1.10 & 4 & 1.10 \\
    UF5 & 0.63 & 0.04 & 1028& 1.10 & 0.06 & 7 &  3 & 4  & 1.10 & 4 & 1.10 \\
    UF6 & 0.64 & 0.00 & 1237& 1.05 & 0.02 & 0 &  9 & 3  & 1.05 & 3 & 1.05 \\
    UF7 & 0.70 & 0.01 & 1262& 1.03 & 0.01 & 0 &  8 & 3  & 1.03 & 4 & 1.03 \\
    UF8 & 0.66 & 0.07 & 936 & 1.14 & 0.04 & 1 &  5 & 4  & 1.14 & 4 & 1.14 \\
    UF9 & 1.08 & 0.03 & 929 & 1.18 & 0.03 & 0 &  6 & 5  & 1.18 & 6 & 1.18 \\
    UF10 & 0.07 & 0.42 & 799& 1.22 & 0.04 & 0 &  4 & 6  & 1.22 & 5 & 1.22 \\
    
    \multicolumn{12}{c}{}\\% \hline
    \rowcolor[gray]{.8}\multicolumn{12}{c}{\textbf{NSGA-II}}\\\hline
    \rowcolor[gray]{.95} & HV & IGD & Nodes & Edges & Shared & Opt.  & Comp. & Max in. & Mean in. & Max out. & Mean out. \\ \hline 
    UF1 & 0.85 & 0.02  & 217 & 0.96 & 0.02 & 0 &  9 & 3 & 0.96 & 3 & 0.96 \\
    UF2 & 0.85 & 0.02  & 217 & 0.93 & 0.00 & 0 & 14 & 2 & 0.93 & 2 & 0.93 \\
    UF3 & 0.41 & 0.23  & 174 & 1.02 & 0.05 & 3 &  1 & 5 & 1.02 & 3 & 1.02 \\
    UF4 & 0.49 & 0.03  & 122 & 0.88 & 0.00 & 0 & 15 & 1 & 0.88 & 1 & 0.88 \\
    UF5 & 0.49 & 0.12  & 206 & 1.02 & 0.09 & 2 & 3  & 4 & 1.02 & 5 & 1.02 \\
    UF6 & 0.62 & 0.01  & 247 & 0.95 & 0.02 & 0 & 13 & 2 & 0.95 & 2 & 0.95 \\
    UF7 & 0.70 & 0.01  & 282 & 0.99 & 0.04 & 0 &  7 & 2 & 0.99 & 3 & 0.99 \\
    UF8 & 0.53 & 0.11  & 171 & 0.99 & 0.08 & 0 &  7 & 2 & 0.99 & 2 & 0.99 \\
    UF9 & 0.93 & 0.10  & 185 & 1.01 & 0.09 & 0 &  4 & 3 & 1.01 & 3 & 1.01 \\
    UF10 & 0.00 & 1.34 & 127 & 0.95 & 0.07 & 0 &  8 & 3 & 0.95 & 3 & 0.95 \\

\end{tabular}
\end{table}

\begin{figure}[htbp]
\centering
\vspace{-1em}
\begin{subfigure}[!t]{1\textwidth}
        	\includegraphics[width=0.8\textwidth]{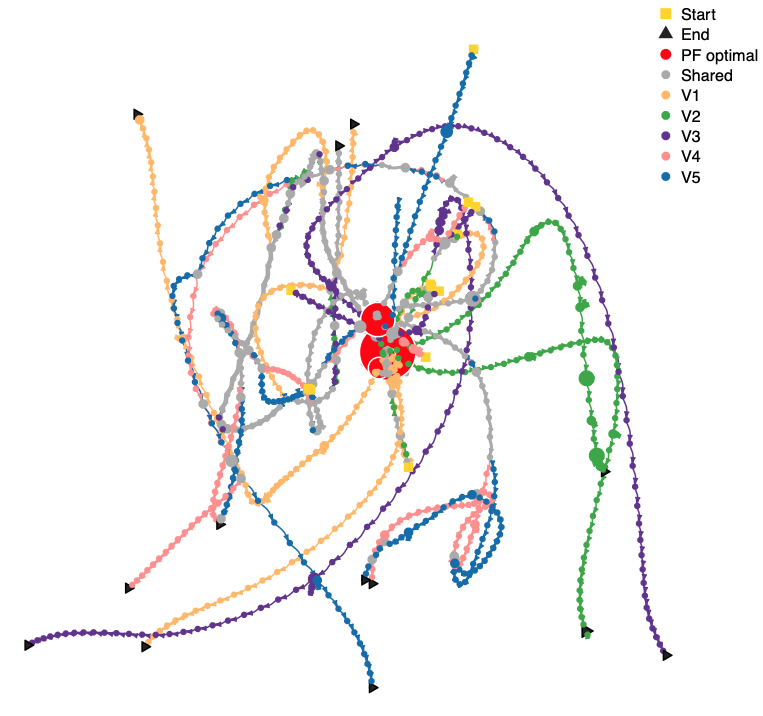}
        % 	\vspace{-5em}
        	\caption{STN of MOEA/D on UF3.}
	\end{subfigure}
	~~
	\begin{subfigure}[!t]{1\textwidth}
        	\includegraphics[width=0.8\textwidth]{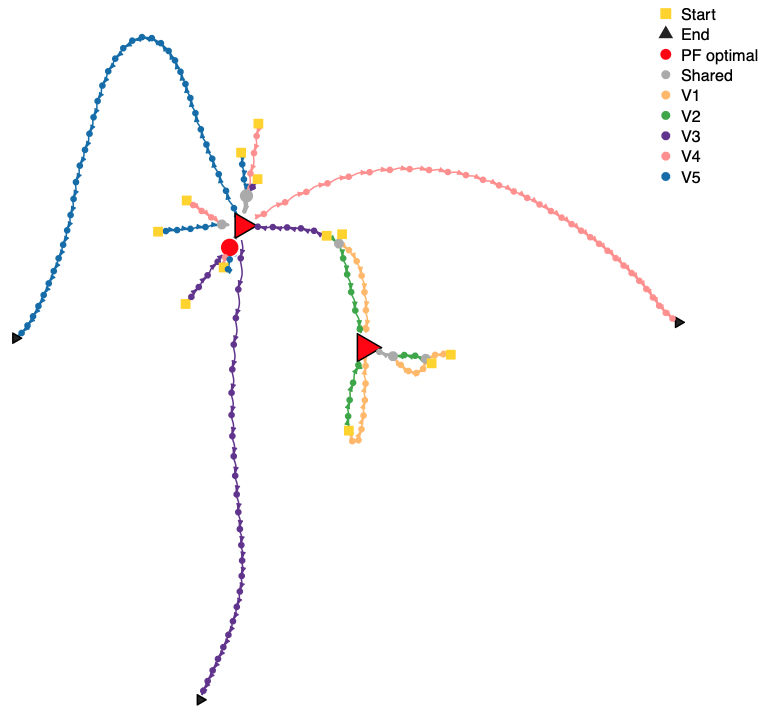}
        % 	\vspace{-5em}
        	\caption{STN of NSGA-II on UF3.}
	\end{subfigure}
\caption{We can see that MOEA/D (top) is able to find optimal solutions during the execution while NSGA-II (bottom) finds optimal solutions mostly at the end of their trajectory (red triangles). This reflects NSGA-II characteristic of using the non-dominance relationship as update criterion.}
\label{fig:stn_UF3}
\end{figure}

\begin{figure}[htbp]
\centering
\vspace{-1em}
\begin{subfigure}[!t]{1\textwidth}
        	\includegraphics[width=0.8\textwidth]{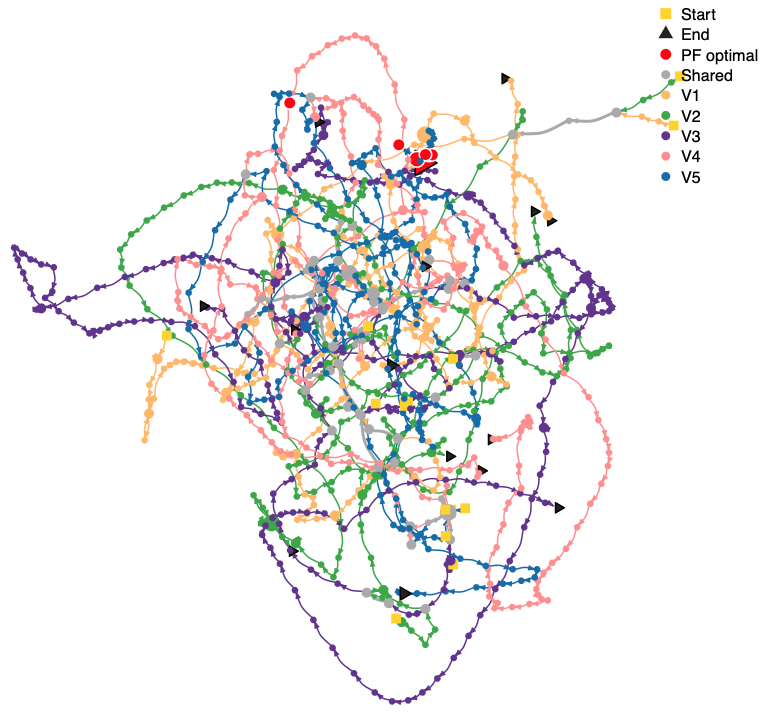}
        	\caption{STN of MOEA/D on UF5.}
	\end{subfigure}
	~~
	\begin{subfigure}[!t]{1\textwidth}
        	\includegraphics[width=0.8\textwidth]{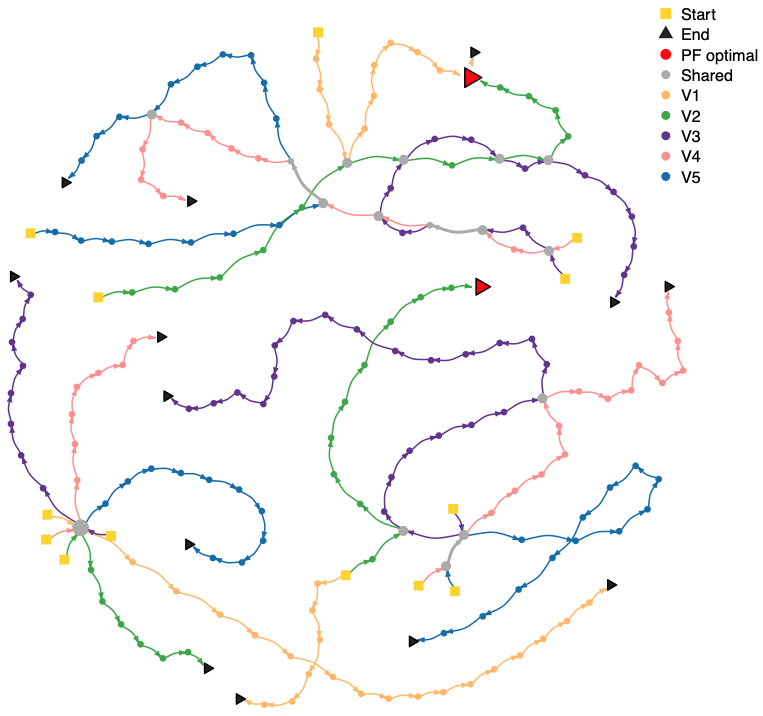}
        	\caption{STN of NSGA-II on UF5.}
	\end{subfigure}
\caption{MOEA/D (top) is able to find many optimal solutions and NSGA-II (bottom) only finds two optimal solutions.}
\label{fig:stn_UF5}
\end{figure}

\begin{figure}[htbp]
\centering
\vspace{-1em}
\begin{subfigure}[!t]{1\textwidth}
        	\includegraphics[width=0.8\textwidth]{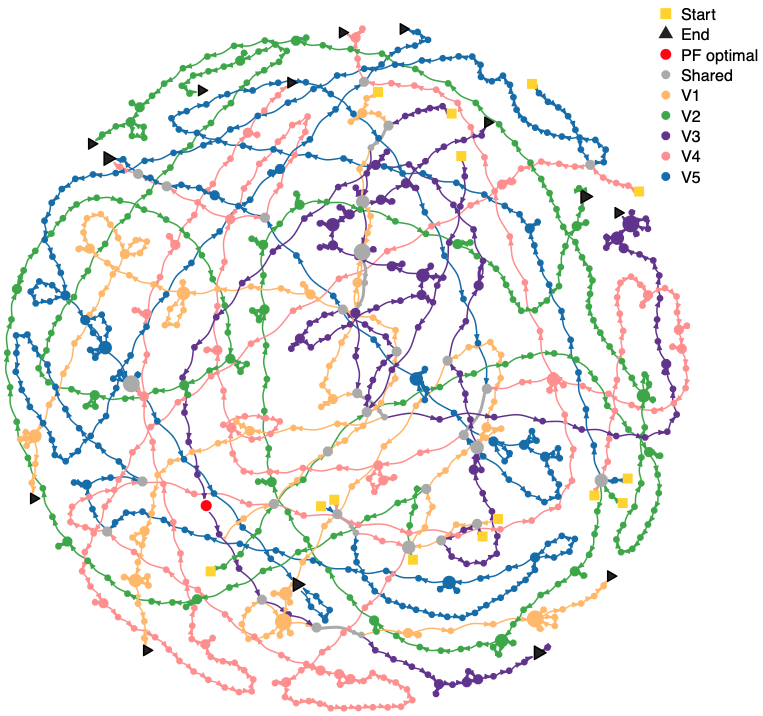}
        	\caption{STN of MOEA/D on UF8.}
	\end{subfigure}
	~~
	\begin{subfigure}[!t]{1\textwidth}
        	\includegraphics[width=0.8\textwidth]{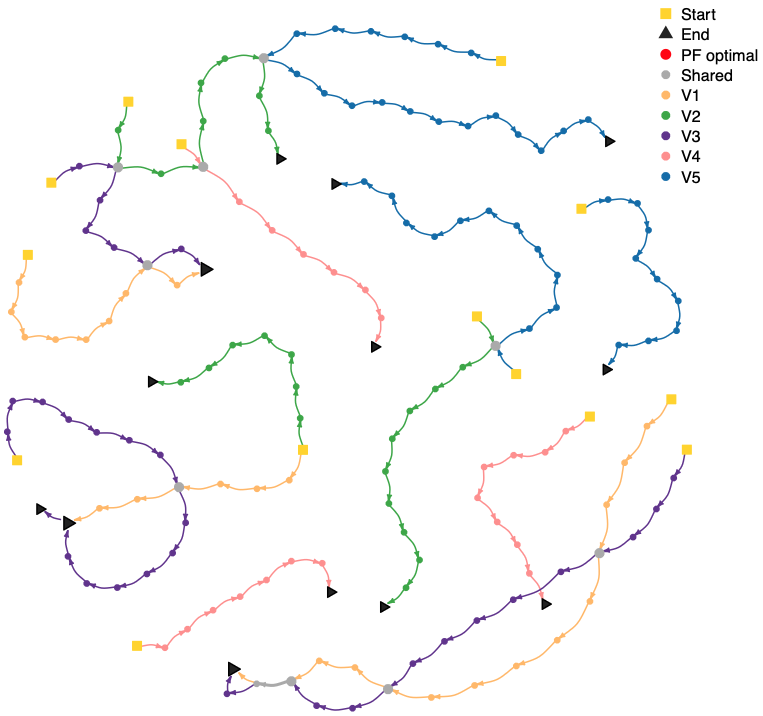}
        	\caption{STN of NSGA-II on UF8.}
	\end{subfigure}
    % \vspace{-1em}
\caption{Only MOEA/D finds optimal solutions. NSGA-II (bottom) progresses the search with almost all vectors sharing nodes, from start to end.}
\label{fig:stn_UF8}
\end{figure}

\begin{figure}[htbp]
\centering
% \vspace{-3em}
	\begin{subfigure}[!t]{0.47\textwidth}
        	\includegraphics[width=1\textwidth]{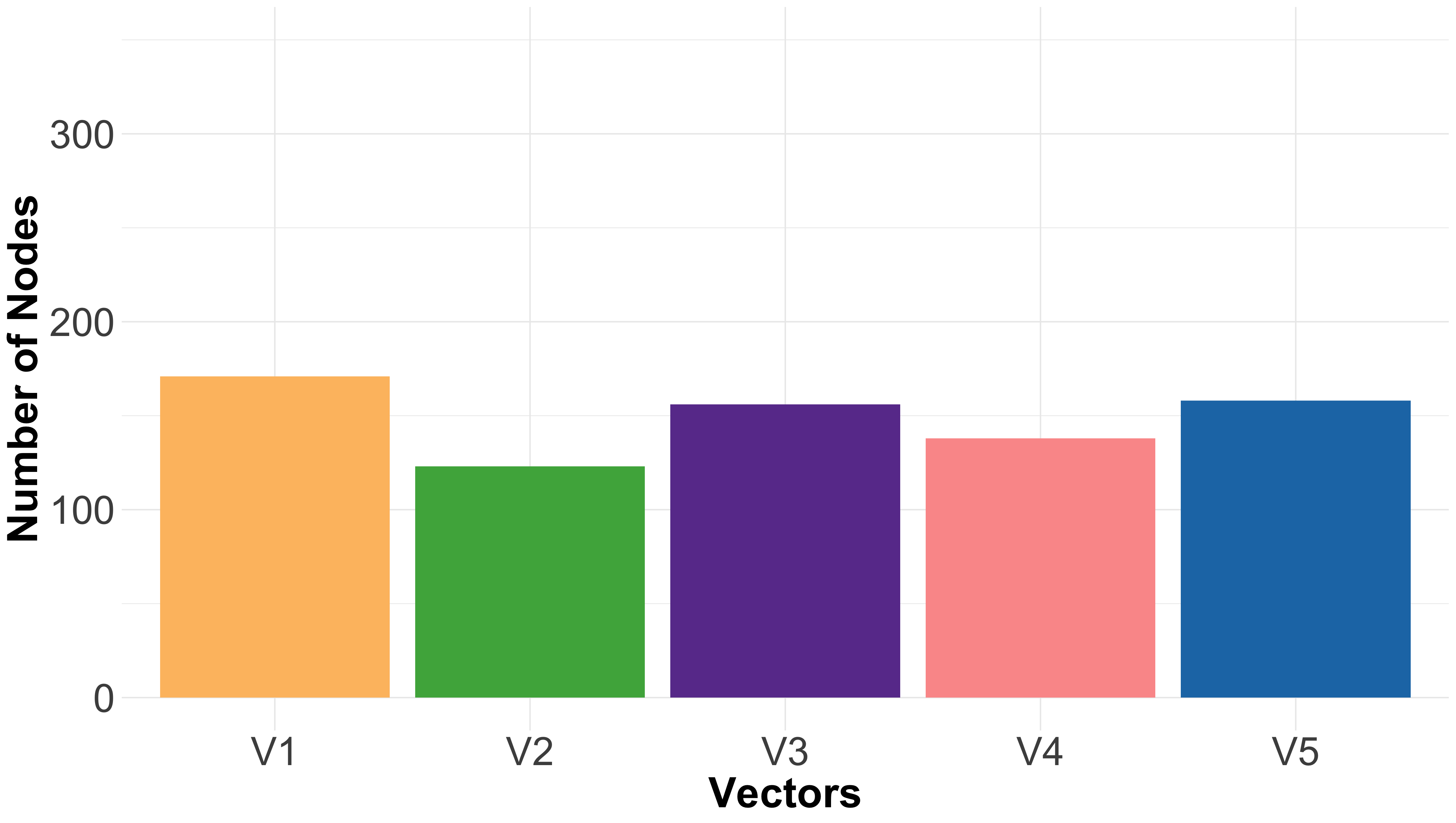}
        	\caption{MOEA/D on UF3.}
	\end{subfigure}
	~~
	\begin{subfigure}[!t]{0.47\textwidth}
        	\includegraphics[width=1\textwidth]{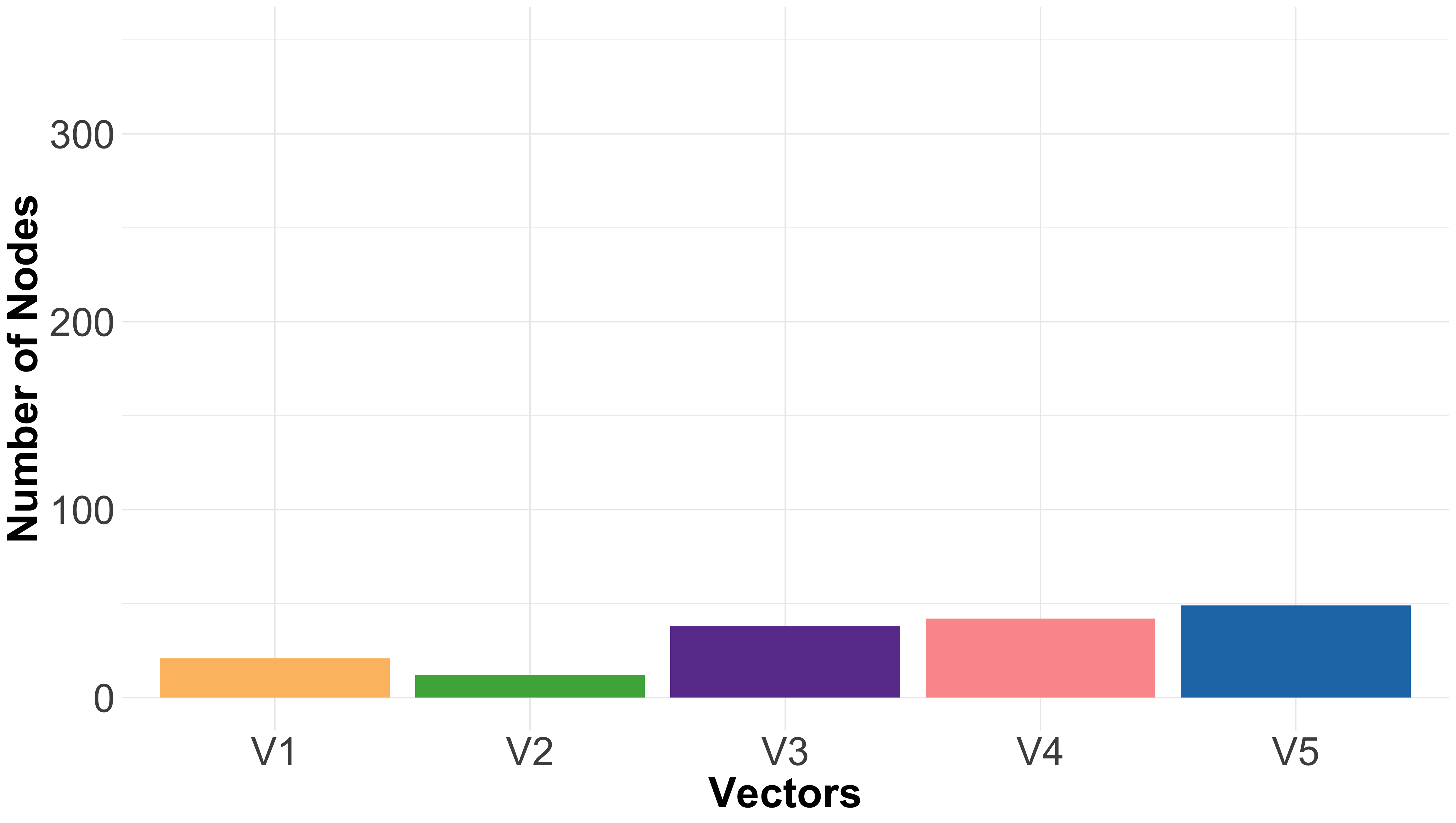}
        	\caption{NSGA-II on UF3.}
	\end{subfigure}
~~
	\begin{subfigure}[!t]{0.47\textwidth}
        	\includegraphics[width=1\textwidth]{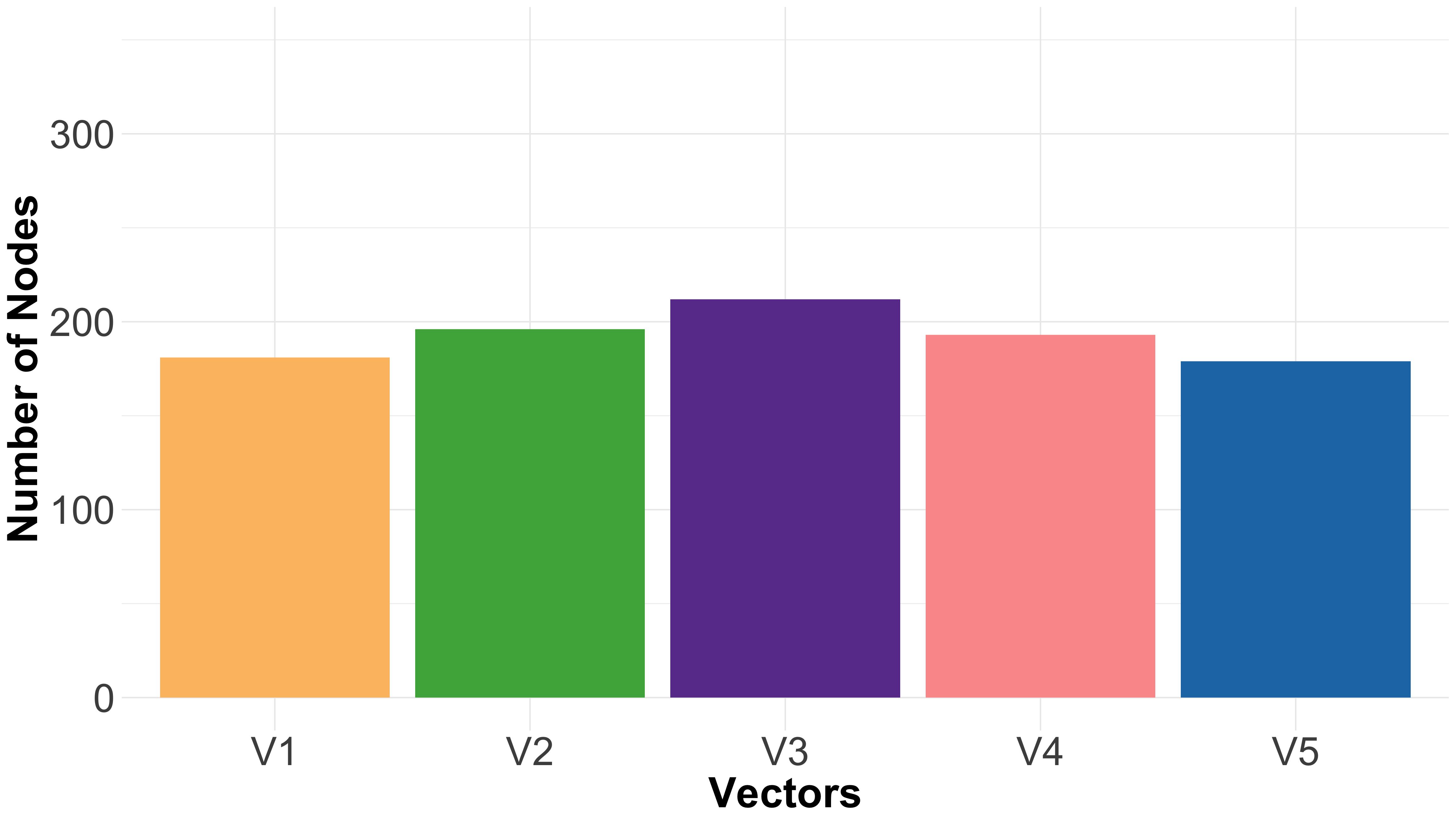}
        	\caption{MOEA/D on UF5.}
	\end{subfigure}
	~~
	\begin{subfigure}[!t]{0.47\textwidth}
        	\includegraphics[width=1\textwidth]{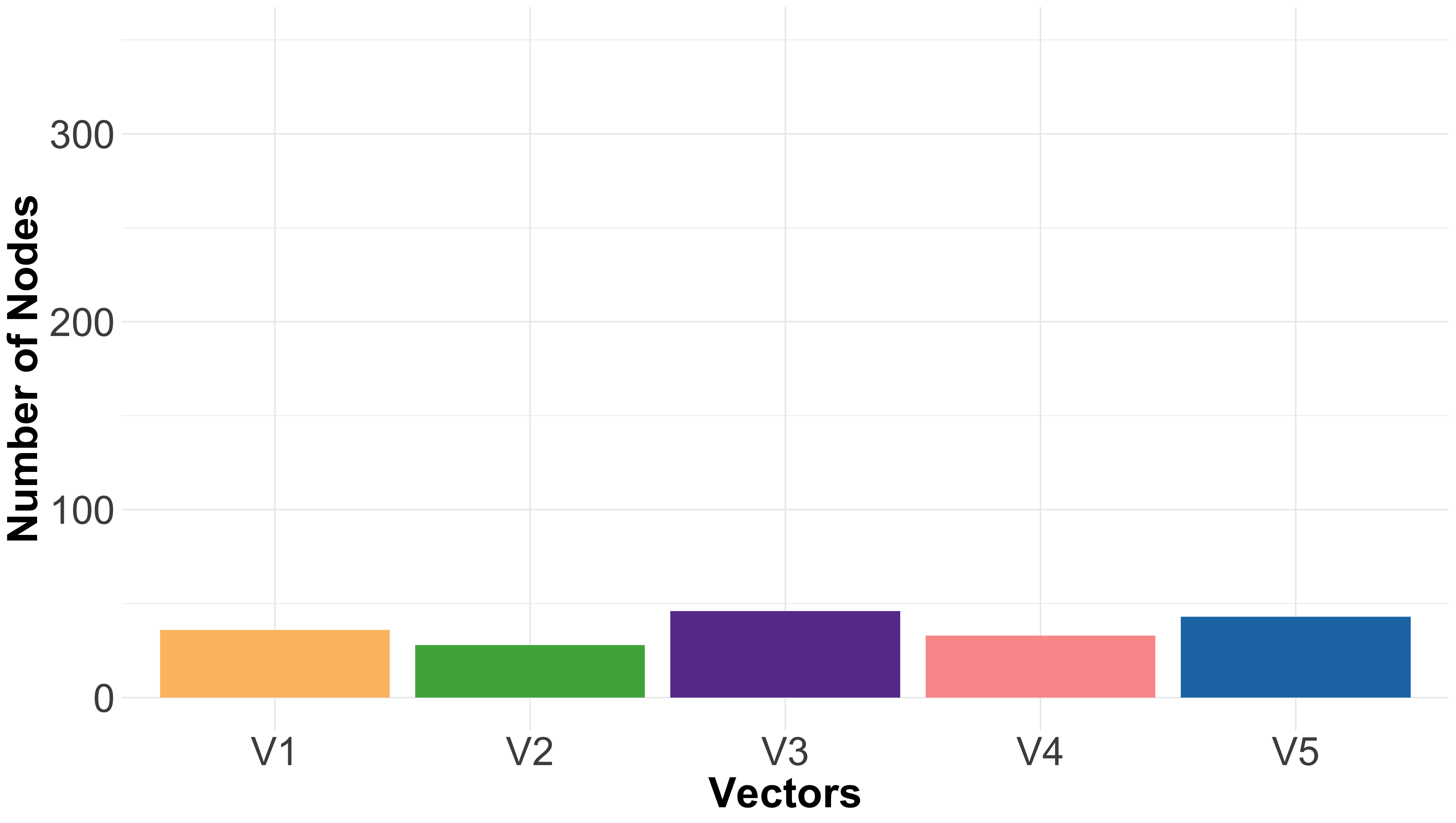}
        	\caption{NSGA-II on UF5.}
	\end{subfigure}
~~
	\begin{subfigure}[!t]{0.47\textwidth}
        	\includegraphics[width=1\textwidth]{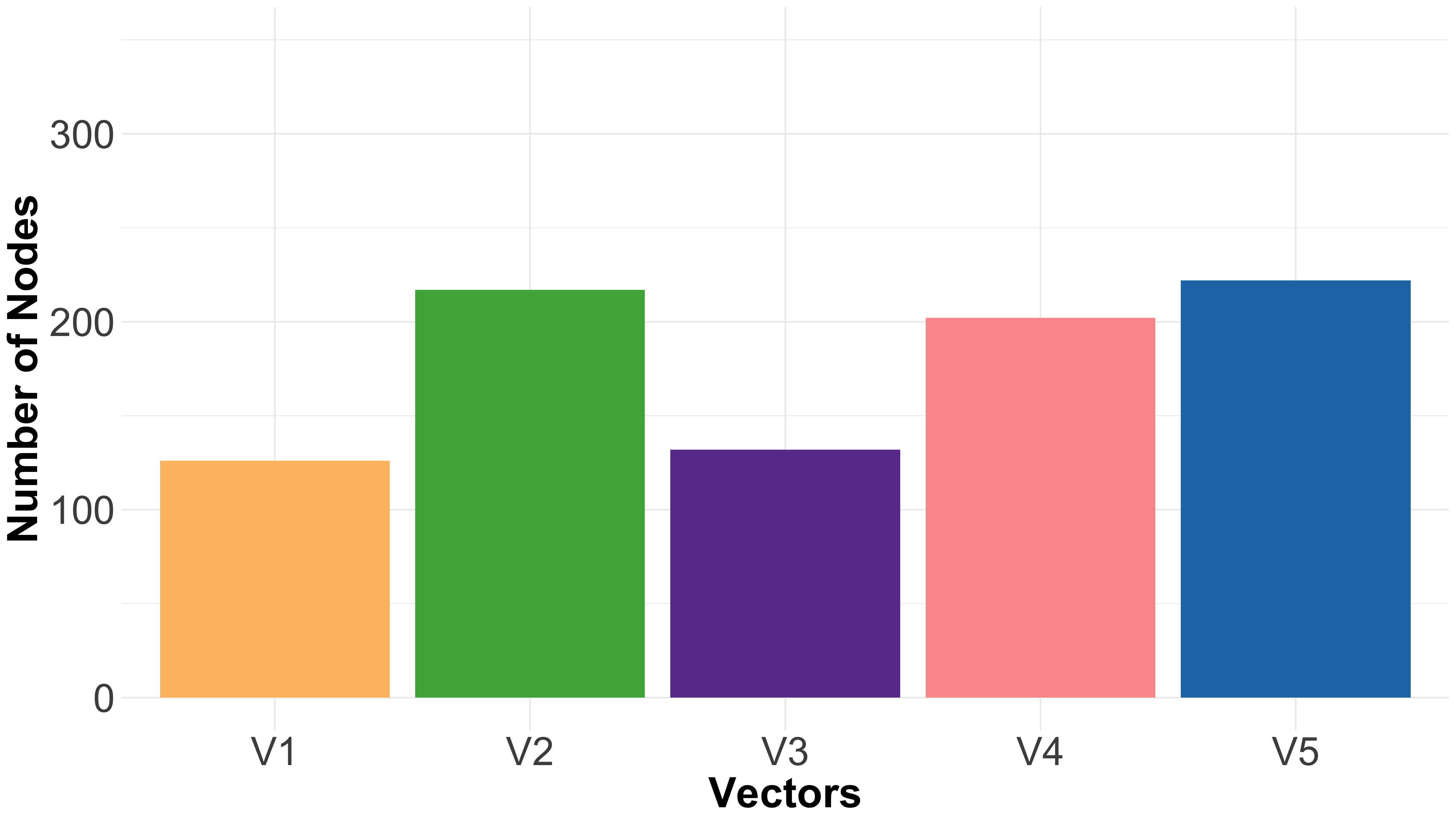}
        	\caption{MOEA/D on UF8.}
	\end{subfigure}
	~~
	\begin{subfigure}[!t]{0.47\textwidth}
        	\includegraphics[width=1\textwidth]{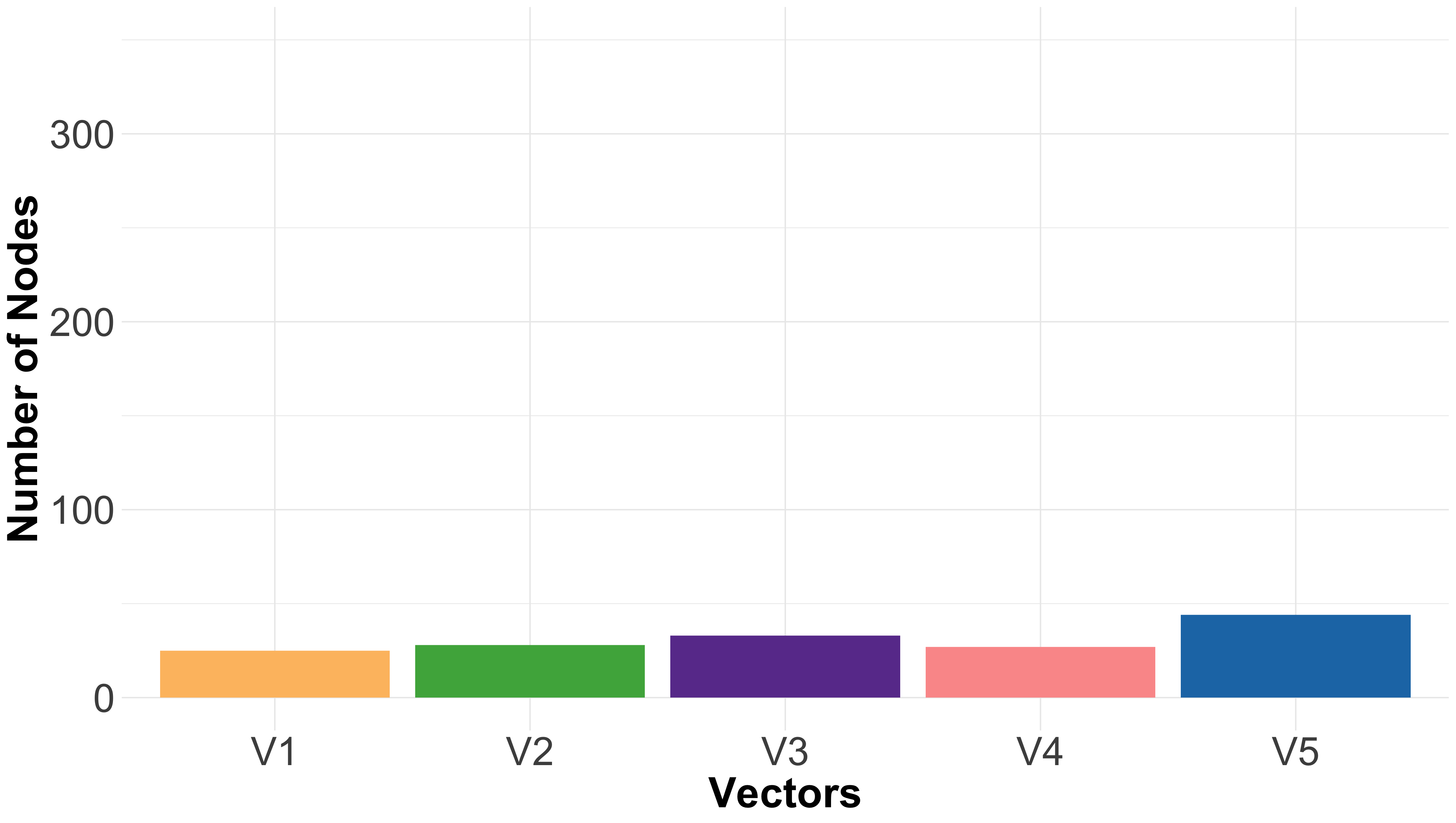}
        	\caption{NSGA-II on UF8.}
	\end{subfigure}
% 	\vspace{-1em}
\caption{The number of nodes for each vector found by MOEA/D is higher than those found by NSGA-II. The distribution of nodes found depends on the problem in question.}
\label{fig:number_nodes}
\end{figure}

This section compares the metaheuristic algorithms MOEA/D and NSGA-II in use of the STN model. For that, we compare the (a) STNs visualizations and (b) the performance in terms of MOP metrics, both HV and IGD, (c) and the STN metrics: (1) the number of nodes and (2) edges, (3) number of shared nodes, (4) the number of Pareto optimal solutions, (5) the number of components of the network, (6) the mean and (7) maximum in-degree and (8) the mean and (9) maximum out-degree.

The visualisations in Figures~\ref{fig:stn_UF3},~\ref{fig:stn_UF5},~\ref{fig:stn_UF8} illustrate the STN by three independent runs of both candidate algorithms on the UF3, UF5 and UF8 benchmark MOPs. Figure~\ref{fig:number_nodes} shows the number
of nodes for each vector traversed by MOEA/D and NSGA-II on these same three MOPs.
The first two MOPs are problems with two objectives, while UF8 has three objectives.  Table~\ref{mop_metrics} displays the traditional HV (higher is better) and IGD (lower is better) metrics and the STN metrics.

Considering the colors used in the STN visualizations, we use yellow squares to indicate the start of trajectories and black triangles to indicate the end of trajectories. The red color shows the best Pareto optimal solutions, and light grey circles show shared locations visited by more than one vector of the same algorithm. Finally, each vector has its color: light orange for V1; green for V2; purple for V3; light pink for V4; and light blue for V5. 

In these visualizations, the size of nodes and the width of edges are proportional to how many times they were visited by the algorithms during the aggregation of runs used to extract the model. Moreover, these visualisations use \emph{force-directed} graph layout algorithms as implemented in \textsf{R} package igraph \cite{Csrdi2006TheIS}, that focuses on position the nodes to have few crossing of edges as possible.

In Figure~\ref{fig:stn_UF3}, we can see the STN modeled for MOEA/D (a, on the top) and NSGA-II (b, on the bottom) on UF3. We highlight that some vector colors do not appear in the images. This is becasue they are following overlapping trajectories and thus are visualized in gray. Looking at the STN for MOEA/D, we can see that all of the Pareto front (PF) optimal solutions, shown in red, are visited multiple times during the search process, as the are large and not at the end of the trajectories. This indicates that MOEA/D can find optimal solutions during the run and continue exploring the decision space. 

On the other hand, the NSGA-II STN  is much smaller, as we can also see in the number of nodes for each vector. It is no surprise that most the PF solutions found by NSGA-II are also at the end of the search since NSGA-II only accepts new solutions if they dominate old ones. Given the size of the red triangles, NSGA-II tends to come back to these solutions frequently. 

Figure~\ref{fig:stn_UF5} shows that the STNs of MOEA/D and NSGA-II are bigger than the STNs modeled for UF3, respectively. We can see that the STN modeled for MOEA/D shows seven optimal solutions. After finding these solutions, MOEA/D continues exploring until the end of the run. Now, NSGA-II also finds two splutions, but tends to come back to them frequently. We can see that the components for both STN have little interaction with which other, suggesting that this problem poses more difficulties to both MOEA/D and NSGA-II than UF3.  

Figure~\ref{fig:stn_UF8} illustrates the UF8 results. We can see that MOEA/D and NSGA-II conduct different searches over the three objective problems (as can also be seen in Table~\ref{mop_metrics}). Looking first at the results of MOEA/D, we can see that this algorithm progresses with the search with a higher exploration rate since, for all vectors, the number of edges and nodes is higher. Also, we can see that the STN of MOEA/D for the UF8 problem is denser than the NSGA-II for the same problem. MOEA/D performs the best in terms of the number of optimal solutions found in a problem, with 1 solution in the theoretical Pareto front. By contrast, NSGA-II finds no optimal solution.

Turning now to the STN metrics for each trajectory, in Figure~\ref{fig:number_nodes}, we can see that each of the MOEA/D vectors visit more nodes than the vectors of NSGA-II, meaning that MOEA/D finds more unique solutions than NSGA-II. This higher number of nodes is not exclusive to the MOPs shown by Figure~\ref{fig:number_nodes}, as can be seen in Table~\ref{mop_metrics}. Moreover, these results, combined with the higher number of optimal solutions reinforce that MOEA/D can explore better the search space given the higher number of optimal solutions. 

In terms of the traditional MOP metrics HV and IGD in Table~\ref{mop_metrics}, we can see that MOEA/D achieves higher HV values and lower IGD values, in comparison with NSGA-II for all MOPs. However, in UF7, both algorithms perform the same, and in UF4, NSGA-II performs the best in both metrics. Overall, for this set of MOPs, we consider that MOEA/D presents better performance than NSGA-II.

Table~\ref{mop_metrics} also shows that the both MOEA/D and NSGA-II algorithm, for most problems, have a ratio of edges around 1. This suggests that the NSGA-II might be exploring small portions of the search space and that MOEA/D is able to explore the search space in depth. That, combined with the lower number of solutions in the theoretical Pareto front, illustrate that NSGA-II faces difficulties in some problems, not exploring many new areas of the search space, and not being able reach the theoretical Pareto front effectively. A closer look at this Table also shows that the maximum out-degree value differs from the in-degree value in UF3 and UF7. This result indicates that these metaheuristics visits the first node more than once. For the UF3 problem we understand that, given the lower number of components and high number of optimal solutions, the metaheuristics are able to find good regions of the search space. The opposite seems to happen for the UF7 problem, since for this problem, there are more components and lower number of optimal solutions.

%summary of the conclusions before
In summary, the STN models suggest an explanation for the differences in HV and IGD performance between MOEA/D and NSGA-II. The STNs showed that MOEA/D extensively explores the search space, finding more optimal solutions without re-visiting nodes too many times. On the other hand, the STNs modeled for NSGA-II indicate that this algorithm faces difficulties when exploring new areas of the search space and that its search progress depends on having good starting points. These results show that using STN for discriminating MOEAs is a viable method for analyzing and aggregating helpful information to the traditional practices for comparing such algorithms, such as the HV and IGD. A note of caution is due here since we only analyzed two MOEAs in a handful set of problems, with only two- and three objectives.

\section{Conclusion}
\label{conclusions}

The main goal of the current study was to determine and design a methodology for applying the search trajectory networks (STNs) for modeling the search behavior of Multi-Objective Evolutionary Algorithms (MOEAs). The proposed method is based on the simple idea of extracting the STN features from some regions of the problem in question. We create STN models in a benchmark set that the most famous MOEAs, MOEA/D and NSGA-II, have different performances. Finally, we show that STNs can be effectively applied to differentiate visually and quantitatively. 

Overall, this study strengthens the idea that characterizing and visualizing the search behavior of MOEAs can provide insightful advances into comprehending how distinct the search behaviors dynamics of such algorithms are and their overall performance. That is because we provided an STN modeling process that can discriminate well different MOEAs, aggregating helpful information to the traditional practices for comparing such algorithms, such as the HV and IGD. Thus, these findings have significant implications for understanding how different MOEAs perform when applied to solve multi-objective problems (MOPs). Moreover, we understand that this work shows excellent results of interest for the whole bio-inspired computation community. 

A natural progression of this work is to investigate the generalization of the model to describe the search behavior of MOEAs other than MOEA/D and NSGA-II, especially in real-world and constrained MOPs. Furthermore, an analysis of the importance of features STN used and introduced in this work should be conducted in the context of automated landscape-aware selection and configuration MOEAs.

\section*{Errata}

The authors found errors in the scripts for constructing and visualising the STNs.  We have updated the code repository, some tables, figures and corresponding text with the corrected STNs.  The general conclusions of the paper remain the same.

% \bibliographystyle{IEEEtran}
% \bibliography{samplebib}

\bibliographystyle{splncs03}
\bibliography{bib}

\end{document}